%% file: main_preprint.tex
\documentclass{article}

\usepackage{arxiv}

\usepackage[utf8]{inputenc} 
\usepackage[T1]{fontenc}    
\usepackage{hyperref}       
\usepackage{url}            
\usepackage{booktabs}       
\usepackage{amsfonts}       
\usepackage{nicefrac}       
\usepackage{microtype}      
\usepackage{xcolor}         
\usepackage{natbib}
\usepackage{graphicx}
\usepackage{amsmath}
\usepackage{algorithm}
\usepackage{algpseudocode}
\usepackage{gensymb}
\usepackage{soul}
\usepackage{adjustbox}

\newenvironment{ack}{%
  \section*{Acknowledgments}%
  \small
}{}

\title{Better Together: Evaluating the Complementarity of Earth Embedding Models}

\author{%
\textbf{Thijs L van der Plas}$^{1}$ \quad \textbf{Jacob JW Bakermans}$^{2}$ \quad \textbf{Vishal Nedungadi}$^1$ \\ \textbf{Gabrielė Tijūnaitytė}$^1$ \quad \textbf{Marc Rußwurm}$^{1,3}$ \quad \textbf{Ioannis N Athanasiadis}$^1$\\
$^1$Wageningen University, NL \quad $^2$University College London, UK \\ $^3$University of Bonn, Germany \\ 
\texttt{\{thijs.vanderplas,ioannis.athanasiadis\}@wur.nl}
}

\renewcommand{\shorttitle}{Better Together: Evaluating the Complementarity of Earth Embedding Models}

\hypersetup{
pdftitle={Better Together: Evaluating the Complementarity of Earth Embedding Models},
pdfsubject={},
pdfauthor={Thijs L van der Plas, Jacob JW Bakermans, Vishal Nedungadi, Gabrielė Tijūnaitytė, Marc Rußwurm, Ioannis N Athanasiadis},
pdfkeywords={},
}

\begin{document}

\input{content}

\end{document}

%% file: content.tex
\maketitle

\begin{abstract}
  Earth embedding models transform Earth observation data into embeddings uniquely tied to locations on the Earth's surface. These models are typically evaluated in isolation, comparing the downstream task performance across different Earth embeddings. However, spatially aligned embeddings can naturally be fused, providing richer information per location, a capability that isolated evaluations fail to capture. We therefore propose assessing Earth embeddings by their \textit{complementarity}: the performance gain of fused embeddings over the best single-model baseline. To operationalise this, we introduce an embedding complementarity index applicable to any embedding and task, and evaluate four Earth embedding models (AlphaEarth, Tessera, GeoCLIP, SatCLIP) in isolation, in all pairs, and jointly across six downstream tasks. Fused embeddings outperform the best single model in four out of six tasks, confirming that single-embedding evaluations often underestimate Earth embedding capabilities. Complementarity proves both task- and location-dependent. Further, for a land cover regression task, we find that complementarity is partially determined by the spatial scale of land cover classes. Complementarity reframes Earth embeddings: the greatest future gains may come not from any single Earth embedding model, but from combinations that are better together.  
\end{abstract}

\section{Introduction}
Earth embedding models construct compressed embeddings from rich, multimodal Earth observation data, which can be used for a variety of geospatial downstream tasks \citep{Klemmer2025earth}. 
Compared to foundation model embeddings from other domains, such as vision or language, Earth embeddings have the unique property that any embedding is inherently linked to a location on the Earth's surface. 
This trait has turned Earth embeddings into commodities, which are pre-computed once and then retrieved by coordinates alone, bypassing the need for model inference \citep{fang2026earth}.
As a result of this ease of access, pre-computed Earth embeddings have now been used for a range of downstream tasks in agriculture, environmental sciences and ecology \citep{zvonkov2025cropland, ma2026harvesting, young2026below, lisaius2026embedding, ball2026geospatial}. But while the \textit{differences} in task performance between Earth embeddings are regularly assessed \citep{zvonkov2025cropland, lisaius2026embedding, ball2026geospatial}, the \textit{complementarity} between different Earth embeddings is not. 

This lack of complementarity assessments might stem from the notion that Earth embedding models are expected to encode all possible information about a location within a single vector. Yet, Earth embedding models are developed with varying training objectives, input data modalities and training datasets. For example, explicit geospatial foundation models map Earth observation images to pixels embeddings, and are trained using masked auto-encoders \citep{brown2025alphaearth} or pixel-wise contrastive loss \citep{tseng2023lightweight, feng2025tessera}, while implicit neural representation models map coordinates to location embeddings via various contrastive loss objectives \citep{vivanco2023geoclip, klemmer2025satclip, Dhakal2025CVPR}. 
This variety of Earth embeddings suggests that each might capture a different ``view" of the same location that is partially unique, and consequently they can be complementary sources of information. To test this, different Earth embeddings can effectively be treated as multiple (embedding) modalities and be fused into multi-Earth embeddings.

In this work, we argue that Earth embeddings should therefore be evaluated both in isolation and in combination, and we demonstrate that for most tasks, Earth embeddings are \textit{better together}. Because Earth embeddings are uniquely indexed by their location, \textit{i.e.}, longitude/latitude coordinates, 
they are spatially aligned and can naturally be fused to obtain one multi-model representation per location. Across six different downstream tasks, we provide a systematic evaluation of four globally available, pre-computed Earth embeddings: AlphaEarth \citep{brown2025alphaearth} and Tessera \citep{feng2025tessera} (pixel embeddings at 10 m resolution), and GeoCLIP \citep{vivanco2023geoclip} and SatCLIP \citep{klemmer2025satclip} (coordinate-based location embeddings). We find that embedding complementarity is task-specific, and we explore the mechanisms that drive embedding complementarity for a land cover regression task. 
In summary, our contributions are:
\begin{itemize}
    \item We propose a complementarity-based evaluation framework and index that quantifies the joint performance gain of fused Earth embeddings over the best single-model baseline.
    \item We evaluate the complementarity of four Earth embedding models across six tasks, demonstrating that complementary embeddings substantially improve performance on four out of six tasks.
    \item We show that embedding complementarity varies spatially and is task-dependent, and that for land cover regression it is partially explained by the spatial scale of land cover classes.
\end{itemize}

\section{Related work}


Multimodal AI models leverage diverse data sources like text, images and depth to perform complex tasks, based off the premise that different modalities can provide complimentary and non-redundant information \citep{alayrac2020self, bachmann2022multimaemultimodalmultitaskmasked, mizrahi20234m, liu2025towards}.  
Similarly, in the natural language processing (NLP) domain, cross-lingual alignment recovers divergent linguistic properties from similar semantic concepts \citep{pallucchini2025lost}. 
Multimodal AI has become integral to geospatial AI, as inherent georeferencing facilitates the integration of diverse Earth observation and geospatial data sources. This has led to the widespread development of Earth embedding models that integrate multiple modalities into embedding vectors \citep{Klemmer2025earth}. However, while Earth embedding model studies routinely quantify the intra-model complementarity of data modalities that feed into Earth embeddings \citep{brown2025alphaearth, jakubik2025terramind}, as well as one-to-one comparisons with other Earth embedding models \citep{feng2025tessera, herzog2025olmoearth, brown2025alphaearth, jakubik2025terramind}, the complementarity between Earth embeddings is rarely assessed. 

Yet, preliminary evidence suggests that combining Earth embeddings can be highly effective. For example, \citet{Dhakal2025CVPR} show that SatMAE pixel embeddings and SatCLIP location embeddings complement each other on a range of geospatial tasks, and \citet{choudhury2026s2vec} assess both one-to-one comparisons and the complementarity of S2Vec with GeoCLIP, SatCLIP and RS-MaMMUT.
Further, location encoders are often used alongside geospatial covariates to boost performance on various geospatial applications, such as species distribution modelling \citep{zbinden2026masksdm}, estimating daily air quality \citep{Karimzadeh2025Performance}, satellite image super-resolution reconstruction \citep{Panangian2025Can}, and land cover/land use prediction \citep{Rao2025Using}. 
Additionally, pixel embeddings can similarly provide additional context that improve predictions on geospatial tasks, including for species distribution modelling \citep{haucke2025seeing} and predicting groundwater contamination \citep{wei2025groundwater}.
Despite these successful examples, a systematic overview that assesses the complementarity of different explicit and implicit Earth embeddings across a range of tasks is missing. Here, we address this by assessing two (explicit) pixel Earth embeddings and two (implicit) location Earth embeddings across six, diverse geospatial tasks. 



\section{Data}\label{sec:data}  
\subsection{Downstream tasks}

We evaluate six downstream geospatial tasks, described below and summarised in Table \ref{tab:overview_tasks}. These tasks encompass a wide range of Earth observation challenges, including \textbf{crop type classification} using the CropHarvest dataset \citep{tseng2021cropharvest}, \textbf{biomass prediction} using MMEarth-Bench \citep{gordon2026mmearth}, \textbf{land cover regression} derived from Dynamic World \citep{brown2022dynamic}, \textbf{bioclimatic variable prediction} derived from WorldClim BIO \citep{odonnell2012bioclimatic}, \textbf{population density prediction} (Pop.) derived from WorldPop (\url{https://www.worldpop.org/}) and \textbf{distance to nearest road prediction} (Dist.\ road) derived from GRIP4 \citep{meijer2018global}. More details regarding the sampling method and preprocessing for each task can be found in Supplementary \ref{downstream_tasks}.

\input{tables/dataset_overview}

\subsection{Retrieving Earth embedding data}
We consider two pixel embedding models (AlphaEarth Satellite Embeddings V1 and Tessera Embeddings V1, both with a 10 m resolution) and two location embedding encoders (GeoCLIP and SatCLIP).

Both AlphaEarth \citep{brown2025alphaearth} and Tessera \citep{feng2025tessera} embeddings are pre-computed per year. Because global Tessera embeddings were only available for 2024 at the time of writing, we retrieved 2024 data for both models, as single-pixel embeddings centred at each location’s coordinate. AlphaEarth, alongside 2024 Dynamic World land cover, bioclimatic, population density and road distance data were downloaded using Google Earth Engine \citep{gorelick2017google}. Tessera embeddings were downloaded from the Tessera API \citep{feng2025tessera}. (Crop type data was downloaded from CropHarvest \citep{tseng2021cropharvest} and biomass data from MMEarth-Bench \citep{gordon2026mmearth}.)

To obtain (time-independent) location embeddings from SatCLIP \citep{klemmer2025satclip} and GeoCLIP \citep{vivanco2023geoclip} we applied the location encoders directly to the sample coordinates.

Embeddings were fused by concatenation to obtain multi-Earth embeddings. For example, AlphaEarth and Tessera embeddings are of sizes 64 and 128, respectively (see Table \ref{tab:overview_models}), meaning that the fused embeddings AlphaEarth + Tessera are of size 192. 


\input{tables/models_overview}

\section{Methods}\label{sec:methods}
\subsection{Downstream task evaluations}
Regression tasks were evaluated using linear regression and the classification task was evaluated using logistic regression. Both methods use L2 sparsity regularisation with $\lambda=1$ and all embedding dimensions and regression target variables were z-scored individually per variable. Datasets were randomly split into 95\% training data and 5\% test data, using 20 folds such that all data points are evaluated as test data exactly once, enabling us to do statistics with $N=20$ folds (see Section \ref{sec:stats}). All statistics reported in this manuscript were calculated from aggregated test set scores only. 

Both univariate and multivariate regression tasks are scored with $R^2$ [\%] and the classification task is scored with accuracy [\%]. For multivariate regression tasks, $R^2$ was averaged across the (equivariant, z-scored) target variables. 

\subsection{Embedding similarity}
We consider two different metrics to quantify the similarity between two embeddings (of different dimensions, see Table \ref{tab:overview_models}): canonical correlation analysis (CCA) and linear centered kernel alignment (CKA) \citep{kornblith2019similarity}. CCA linearly reprojects embeddings to a set of orthogonal vectors (to a space with the same number of dimensions as the smallest embedding) such that the correlations between the projections of two embeddings is maximised. We then report the mean CCA correlation $\rho_{m_1, m_2}^{CCA}$. Linear CKA was originally designed for cases where the number of features outweighs the number of samples, and can be thought of as CCA weighted by its Eigenvalues \citep{kornblith2019similarity}. Hence, it assigns greater weight to projection vectors (and the correlations between them) that explain a greater proportion of the variance.

\subsection{Embedding complementarity}

We define a complementarity index $C$ for two models $m_1$ and $m_2$ on task $t$ as:
\begin{equation}\label{eq:compl_2_models}
    C^t_{m_1, m_2} = \frac{S^t_{m_1 + m_2} - \text{best} \left( S^t_{m_1}, S^t_{m_2} \right)}{S^t_{\text{best}} - \text{best} \left( S^t_{m_1}, S^t_{m_2} \right)} \leq 1
\end{equation}
where $S_m^t$ is the score of model $m$ on task $t$, and $S_{\text{best}}^t$ is the best possible score (\textit{e.g.}, 1 for $R^2$ or 0 for mean squared error (MSE)). 
This index calculates the complementarity with respect to the best-performing model, and is $0 < C^t_{m_1, m_2} \leq 1$ whenever the multi-Earth embedding performs better and $C^t_{m_1, m_2} \leq 0$ when it performs worse, regardless of whether the original metric should be minimised or maximised. For more than two models, we extend the definition of this index to:
\begin{equation}\label{eq:compl_n_models}
    C^t_{m_1, m_2, \ldots, m_n} = \frac{S^t_{m_1 + m_2 + \ldots + m_n} - \text{best} \left( S^t_{m_1}, S^t_{m_2}, \dots, S^t_{m_n} \right)}{S^t_{\text{best}} - \text{best} \left( S^t_{m_1}, S^t_{m_2}, \dots, S^t_{m_n} \right)} \leq 1
\end{equation}

\subsection{Spatial scale}
We estimated the spatial scale $d$ for each land cover class $y_i$ (Fig \ref{fig:lc_maps}), defining it as the exponential decay coefficient of the difference in land cover class probabilities between locations. This metric quantifies the spatial continuity of each land cover class. Our approach of calculating spatial scale is conceptually similar to the calculation of the effective range using variograms \citep{vanderplas2025neureo}, but with the modifications: we use entropy instead of variance, and we weigh pairs by their maximum land cover probability to filter $(0, 0)$ pairs.
In more detail, the spatial scale per land cover class is calculated as follows:

First, we calculated the L2 norm of the land cover probabilities for every pair of locations $(y_{i,l_1} - y_{i, l_2})^2$, and plotted this versus the Haversine distance between the locations, for all points with distances $\leq 1000$ km. We then calculated the density of this plot with 100 bins along each axis, where we weighed each point by the maximum land cover value $\max{(y_{i,l_1}, y_{i, l_2})}$ (Fig \ref{fig:spatial_entropy_lc}, bottom row). (Data points were weighed because land cover probabilities are normalised across the 9 classes, meaning that for any class the majority of $y_i \approx 0$, which, when not corrected for, would heavily skew the density function to $(y_{i,l_1} - y_{i, l_2})^2 = 0$.) 
Finally, to quantify how quickly the distribution of L2 differences between locations widens as a function of distance, we calculated the entropy per distance bin and fit this curve with an exponential fit $H(x) =H_\text{max} - (H_\text{max} - H(0)) \cdot \exp (-\frac{x}{d})$ (Fig \ref{fig:spatial_entropy_lc}, top row). We fit each land cover class separately, and define the exponential decay coefficient $d$ as its spatial scale.

\subsection{Statistics}\label{sec:stats}
Average test scores across folds are reported as mean $\pm$ standard error of the mean (SEM). 
We used one-sided Wilcoxon signed-rank tests to test the statistical significance of test set scores across folds ($R^2$ for regression, accuracy for classification) \citep{rainio2024evaluation}. Exact p-values $p$ are reported in the text. In tables, significance is indicated by asterisks (*: $p<0.05$, **: $p<0.01$, ***: $p<0.001$), representing p-values adjusted for multiple comparisons using the Benjamini-Hochberg method with a false discovery rate of 0.05.



\subsection{Software and code}\label{sec:software_code}

All our code is available open-source at \url{https://github.com/vdplasthijs/better_together}. All datasets required to reproduce the results in this paper are available as csv files in the same repository under a CC-BY-SA 4.0 license.
All experiments were performed on a Macbook Pro M4 Pro 24 GB. Downstream task evaluations and analyses took seconds each, meaning that all experiments can be completed in under 10 minutes of execution time. 

\section{Results}
\subsection{Task-agnostic similarity and task complementarity}
\begin{figure}[t]
    \centering
    \includegraphics[width=\linewidth]{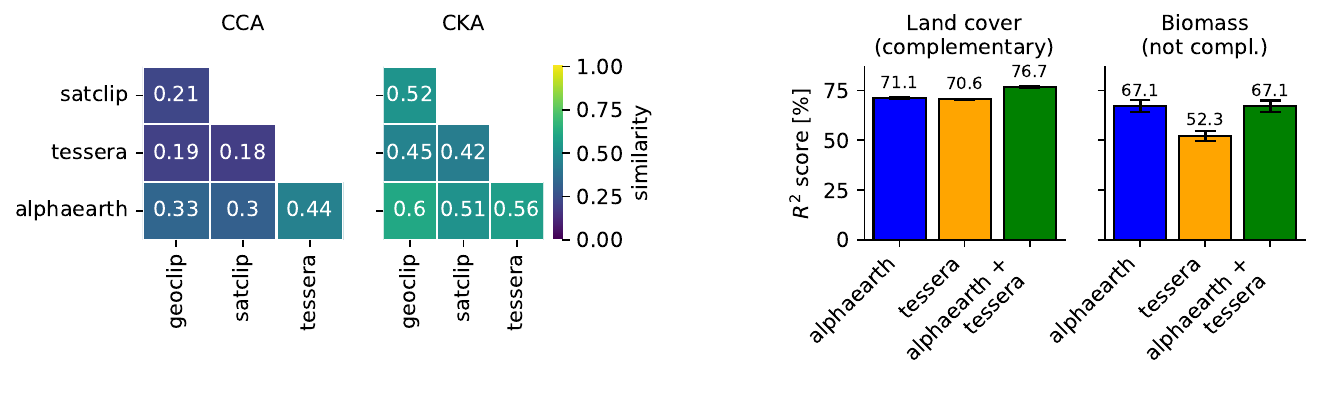}
    \caption{\textbf{Embeddings are partially similar and partially complementary.} \textbf{a)} CCA and CKA similarity values for each pairwise combination of four Earth embeddings indicate that all embedding combinations are substantially, but not completely, similar. \textbf{b)} Embedding complementarity was assessed by comparing their fused downstream task performance (green) against their individual task performances (blue, yellow). In this example, AlphaEarth and Tessera are complementary for the land cover task, but not for the biomass task. Error bars denote SEM across test splits.}
    \label{fig:similarity_compl_example}
\end{figure}

Even though each Earth embedding model encodes geospatial information differently, their embeddings are correlated. We quantified this for four different Earth embedding models, using a set of 8,901 locations, distributed spatially uniformly across the Earth's landmass, quantified by CCA and CKA similarity (Fig \ref{fig:similarity_compl_example}a). 
We observe CCA similarity values between 0.18 to 0.44, and significantly higher similarity values using CKA (0.42 to 0.60, $p = 2 \cdot 10^{-4}$, one-sided t-test, Fig \ref{fig:similarity_compl_example}a). Compared to CCA, CKA indicates that the dominant, low-dimensional modes of embeddings are especially correlated, for all Earth embedding pairs we tested. 


At the same time, no two Earth embeddings are (close to) 100\% correlated, meaning that different embeddings can potentially complement each other by providing different pieces of information about the same set of locations. To establish that these decorrelated parts are not just noise, we evaluated the complementarity of Earth embeddings by comparing the predictive performance on downstream tasks of fused embeddings with the performance of single-Earth embeddings. As an example, AlphaEarth and Tessera are complementary on the land cover regression task, because their fused score (76.7\%) is significantly greater than any of their individual scores (71.1\% and 70.6\%, respectively) ($p=9.5 \cdot 10^{-7}$, one-sided Wilcoxon signed-rank test, Fig \ref{fig:similarity_compl_example}b). However, they are not complementary on the aboveground biomass prediction task where their fused score is not significantly different from their best individual score ($p=0.58$, one-sided Wilcoxon signed-rank test, Fig \ref{fig:similarity_compl_example}b). Hence, their complementarity depends on the downstream task, which we further quantify in the next section by evaluating all combinations of four Earth embedding models on six tasks. 

\subsection{Embedding complementarity is task-dependent}
We find that for four out of six downstream tasks, fusions of pairwise combinations of four different Earth embeddings beat the best single-Earth embedding per task (Table \ref{tab:r2_scores_main_tasks}). For each of these tasks, the best embedding pair includes the best single-Earth embedding. However, we also find that for three out of these four tasks, an embedding pair that excludes the best single-Earth embedding can rival or outperform the best single-Earth embedding. 
Further, fusing all four Earth embeddings together beat the best pair of embeddings for only two out of six tasks, showing that the complementarity between embeddings saturates as more embeddings are added. 

\input{tables/r2_scores_main_tasks}

\begin{figure}
    \centering
    \includegraphics[width=0.8\linewidth]{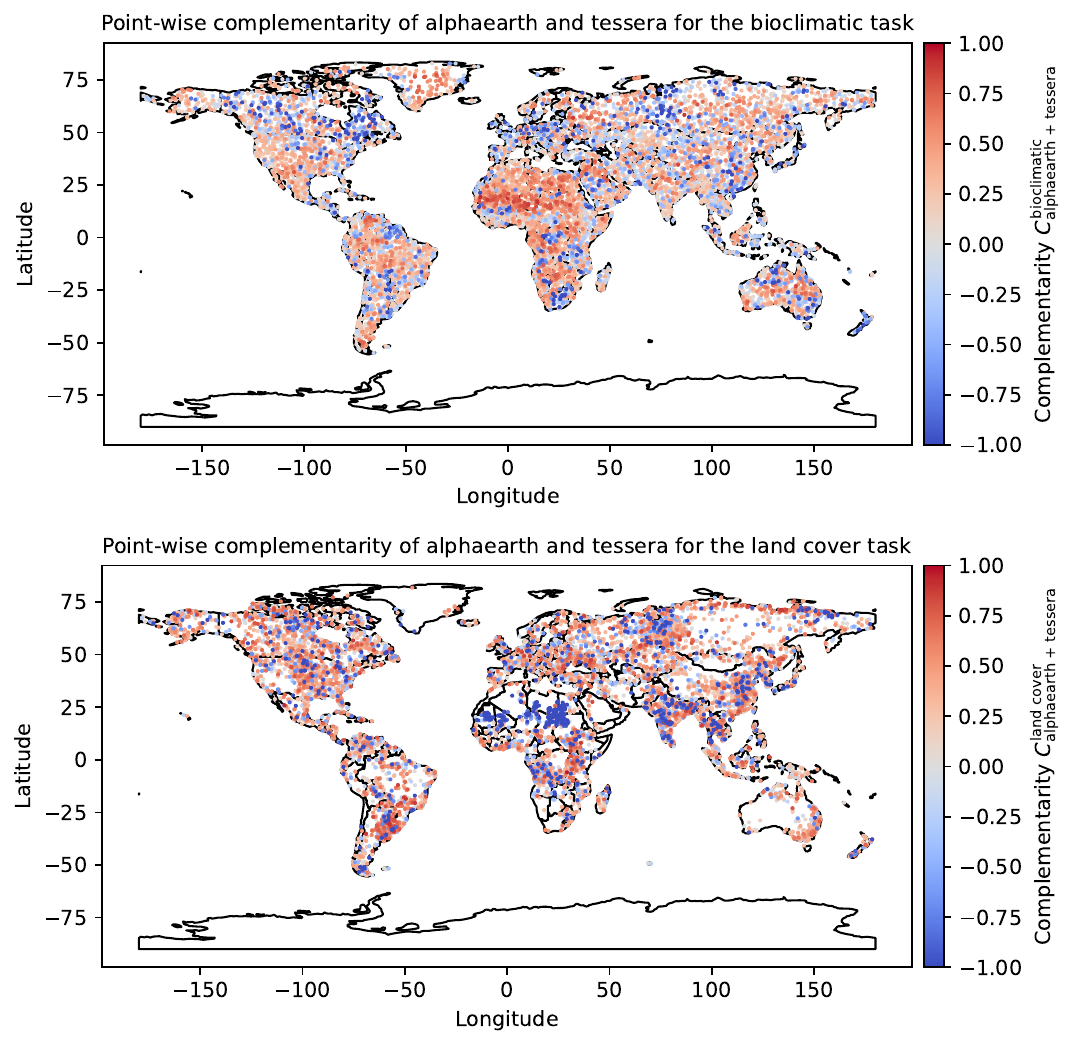}
    \caption{\textbf{Embedding complementarity varies spatially.} Complementarity was computed per point for AlphaEarth and Tessera, for the bioclimatic (top) and land cover (bottom) tasks, using the MSE per point. Positive values indicate positive complementarity, and negative values were clipped at $-1$.}
    \label{fig:example_location_compl}
\end{figure}

We find that embedding complementarity not only strongly depends on the downstream task (Table \ref{tab:compl_r2_scores_main_tasks}), but also varies spatially, as determined by calculating the complementarity index (Eq \ref{eq:compl_2_models}, \ref{eq:compl_n_models}) per location using the location-wise MSE (Fig \ref{fig:example_location_compl}). Notably, these spatial dependencies exhibit a task-location co-dependency. The geographic distribution of complementarity strength is not static, but rather shifts according to the specific task domain (Fig \ref{fig:example_location_compl}).

\input{tables/compl_r2_scores_main_tasks}

\subsection{Land cover spatial scale explains SatCLIP complementarity}

The location embeddings of SatCLIP complement the pixel embeddings AlphaEarth and Tessera on the land cover task, even though SatCLIP scores much lower individually (Table  \ref{tab:r2_scores_main_tasks}, \ref{tab:compl_r2_scores_main_tasks}). This low average score of SatCLIP is caused by strongly varying performance per land cover class: evaluations on individual land cover class regression show scores ranging from $7.6 \pm 1.4$ on \textit{water} to $80.1 \pm 0.7$ on \textit{snow and ice} (Table \ref{tab:r2_scores_lc_classes}). 
However, despite low scores on most classes, SatCLIP significantly complements both AlphaEarth and Tessera on almost all land cover classes (Table \ref{tab:compl_r2_scores_lc_classes}). Further, GeoCLIP significantly complements AlphaEarth on 4 classes and Tessera on 1 class.

The different land cover classes are evaluated for the same locations, therefore differences in performance may result from class-specific spatial covariance (see Fig \ref{fig:lc_maps}). These differences are expressed by the spatial scale $d$ values of each land cover class, which represent at what scales land cover classes become dissimilar (Fig \ref{fig:spatial_entropy_lc}). We find $d$ values ranging from 42 km (\textit{flooded vegetation}) to 581 km (\textit{bare}) (Fig \ref{fig:spatial_entropy_lc}), highlighting the different profiles of land cover distributions (Fig \ref{fig:lc_maps}). The spatial scale only correlates significantly with SatCLIP (Spearman $\rho=0.80$, corrected $p=0.02$) and GeoCLIP (Spearman $\rho=0.68$, corrected $p=0.04$) performance, and only significantly correlates with complementarity for embedding combinations that include either SatCLIP or GeoCLIP (all such combinations except AlphaEarth + GeoCLIP, see Table \ref{tab:spatial_r2}, \ref{tab:spatial_compl}). 

\input{tables/r2_scores_lc_classes}

\input{tables/compl_r2_scores_lc_classes}

\begin{table}[h]
    \centering
    \begin{minipage}{0.48\textwidth}
        \centering
        \input{tables/spatial_relationship_r2}
    \end{minipage}
    \hfill
    \begin{minipage}{0.48\textwidth}
        \centering
        \input{tables/spatial_relationship_compl}
    \end{minipage}
\end{table}

\section{Limitations}\label{sec:limitations}
In this study, we evaluated the complementarity of four Earth embedding models (which have pre-computed embeddings available) on six downstream tasks. As many more Earth embedding models exist, as well as a greater variety of downstream tasks, this is not a comprehensive overview that aims to conclusively determine the complementarity between all Earth embedding models. Rather, our results demonstrate that, for the majority of model combinations and tasks we assessed, embedding complementarity exists and can improve task performance. We hope that this motivates future Earth embedding model research to include complementarity tests in task evaluations.

Location embeddings vary at greater scales than pixel embeddings. While pixel embeddings are calculated directly from high-resolution satellite imagery, and thereby capture fine-grained local features, location embeddings are calculated directly from geographic coordinates, which produce smoother representations. Because we sought to compare location embeddings directly with pixel embeddings, we chose to extract one pixel embedding value per location (to match the single location embedding per location), and we focused on the evaluation of the embeddings on global tasks. These two constraints mean that we do not utilise the full capabilities of pixel embeddings. For example, pixel embeddings also excel at high-frequency tasks such as semantic segmentation \citep{brown2025alphaearth, feng2025tessera}, a domain where location embeddings struggle due to their inability to resolve spatial variability at the pixel scale (of $\sim10$ m). Alternatively, pixel embeddings and location embeddings could be compared for semantic segmentation task by using the pixel embeddings to perform semantic segmentation, while using the location embedding as a fixed context per patch (\textit{e.g.}, see \citet{Rao2025Using}). 

We found that for two downstream tasks, fusing all embeddings was best, while for three tasks fusing two embeddings was best, and for two tasks a single-embedding model performed best. In theory, fusing embeddings (by concatenation) should not decrease their potential task score, as a trivial solution would be the single-embedding solution concatenated with zero weights. However, in practice adding covariates that are essentially noise or redundant can limit performance due to overfitting (Occam's razor). All tasks have between 3.5k and 9k data points, while concatenating all embeddings yields 960 fused embedding dimensions.
Although specialized prediction heads (such as those incorporating bottleneck layers) could be optimized to mitigate overfitting, our results highlight that identifying the most complementary combination of embeddings can be a more efficient path to improve performance.
However, further compression of fused embeddings is likely to be beneficial given their positive similarity, hence redundancy (Fig \ref{fig:similarity_compl_example}). An exciting future research direction would be identify the optimal task- and location-specific compression of fused Earth embeddings.

\section{Discussion}
Our results show that as Earth embedding models learn to represent the same data point (\textit{i.e.}, location), differences in their architectures and training strategies can yield embeddings that capture (partially) different aspects of the data. We demonstrated that this complementary property can be leveraged by fusing the embeddings, leading to better task performance. However, theoretically, the information encoded by one Earth embedding might be a subset of the information encoded by another Earth embedding, and complementarity might no longer exist. This would align with the Platonic Representation Hypothesis, which argues that as models become more capable, their embeddings converge towards a shared latent structure of reality \citep{huh2024platonicrepresentationhypothesis}. Under this formulation, different encoders on similar data are not arbitrary, but are different projections of the same underlying structure. While, some recent evidence supports convergence towards a shared local latent structure \citep{groger2026revisiting}, others argue that the convergence may be domain-dependent or hindered by architectural biases \citep{koepke2026back}. 
We observed diminished returns as more Earth embeddings were fused: only for two out of six tasks did the combination of all four Earth embeddings outperform the best pair. However, these fused embeddings did not reach the best possible accuracy (of 100\%), meaning that even though their embeddings may be converging onto a shared information space, this information is not sufficient to solve the task (assuming the target variables of our tasks are noise-free). Other Earth embedding models, using different data modalities, training strategies, or model architectures, could hence further complement the embedding space.

Therefore, we argue that whether Earth embedding models, or foundation models more generally, will eventually reach a singular "Platonic" state is less critical than their complementarity on the path toward it. Developing multiple models, using different architectures, and fusing their embeddings, might prove a more efficient strategy in practice to obtain a comprehensive embedding space. A remaining challenge (for Earth embedding models) is to compress a fused embedding space, either independently of or specific to downstream tasks.

\section{Conclusion and outlook}
Despite its importance, assessing the true complementarity of different modalities remains a key challenge in the development of multimodal AI systems \citep{liu2025towards}. We found that by evaluating the complementarity of four different Earth embedding models, fusing Earth embeddings improved the performance for four out of six representative geospatial tasks. Complementarity was strongly task-dependent, with different Earth embedding combinations yielding the highest complementarity for different tasks (Table \ref{tab:compl_r2_scores_main_tasks}). We attempted to better understand this task-dependence by further inspecting the land cover task, and found that here, complementarity is partially governed by the inherent spatial scales of both models and tasks: the coarser-scale SatCLIP and GeoCLIP embeddings contribute strongest to tasks that vary on longer ranges. Future research should aim to understand the relation between model complementarity and task more comprehensively, which is needed to efficiently find complementary Earth embedding combinations for geospatial applications.

Based on our results, we propose that Earth embeddings should not only be compared against, but also together with, other Earth embeddings. Further, as a consequence of allowing embeddings to be fused, we argue that future Earth embedding model developments should focus on providing information that is complementary to existing, pre-computed embeddings. 


\begin{ack}
TLvdP and INA acknowledge funding from the Dutch Research Council (NWO) Large-Scale Research Infrastructures (LSRI) programme for the LTER-LIFE (\url{http://www.lter-life.nl}) infrastructure (grant 184.036.014). JJWB acknowledges funding from the Royal Society through a Royal Society Newton International Fellowship (NIF\textbackslash R1\textbackslash 252778). VN acknowledges support from the Digital Europe Programme under Grant agreement \href{http://www.agrifoodtef.eu}{AgrifoodTEF} - Test and Experiment Facilities for the Agri-Food Domain (Grant \#\href{https://ec.europa.eu/info/funding-tenders/opportunities/portal/screen/opportunities/projects-details/43152860/101100622}{101100622}). MR acknowledges funding by the Deutsche Forschungsgemeinschaft (DFG, German Research Foundation) – project number 572735710.
\end{ack}
\clearpage 

\bibliographystyle{plainnat}
\bibliography{bibliography}

\clearpage
\appendix
\renewcommand{\thefigure}{S\arabic{figure}}
\renewcommand{\thetable}{S\arabic{table}}
\setcounter{figure}{0}
\setcounter{table}{0}

\section{Technical appendices and supplementary material}\label{sec:supp_mat}

\subsection{Downstream tasks}
\label{downstream_tasks}
We provide more information below on the downstream tasks used for the evaluation. For each task, we first note the initial number of locations. For our study, we retained locations only if their embeddings are available across all models. This filtering resulted in slightly reduced location counts per task, as denoted in Table \ref{tab:overview_tasks}.

\paragraph{Crop type} Crop type classification task of 18 classes, derived from the CropHarvest dataset \citep{tseng2021cropharvest}. We used the global CropHarvest dataset and selected all classes with $\geq200$ samples, yielding 18 out of 306 classes. We then randomly subsampled all 18 classes to 200 samples each to obtain a balanced dataset of 3,600 locations. 

\paragraph{Biomass} Biomass prediction task, using the MMEarth-Bench biomass dataset \citep{gordon2026mmearth}. The original dataset has 18,393 locations, each containing a 128 $\times$ 128 pixel patch at 10 m resolution with sparse aboveground biomass measurements. We selected the patches that did not have a \texttt{NaN} value at their centre-pixel, resulting in 4,585 locations. We considered two tasks: predicting the centre-pixel biomass value and predicting the average (non-\texttt{NaN}) biomass values of the patch. $R^2$ scores were higher for the average biomass task, and differences between embeddings were comparable across the two tasks. We therefore included only the average biomass task in the main results, but provide the results of centre-pixel task in the Supplementary materials (Table \ref{tab:r2_scores_main_tasks_biomass}). 

\paragraph{Land cover} Land cover multi-label regression task of 9 classes, using the Dynamic World dataset \citep{brown2022dynamic}. Dynamic World predicts land cover across 9 classes for each acquired Sentinel-2 satellite image, effectively yielding a predicted land cover time-series. We retrieved the Dynamic World land cover data for the year 2024, averaged across all 2024 data points, yielding a probability between 0 and 1 for each land cover class (normalised for each location). We created a balanced dataset of 10,000 locations, by sampling locations globally but stratified across the 9 classes (see Supplementary Section \ref{sec:sampling_lc_locations} for further details, all classes were balanced except for \textit{flooded vegetation}). This set of locations was also used for the population density task, described below, because it oversamples urban areas compared to spatially uniform sampling. 

\paragraph{Bioclimatic} Prediction task of 19 bioclimatic variables, using the WorldClim BIO dataset \citep{odonnell2012bioclimatic}. We created a dataset of 10,000 locations sampled randomly and spatially uniformly across the Earth's landmass (excluding Antarctica), and retrieved the WorldClim BIO data for these locations. This set of locations was also used for the distance-to-nearest-road task, described below. 

\paragraph{Population density} Prediction task of population density (per km$^2$), using the WorldPop dataset (\url{https://www.worldpop.org/}). 

\paragraph{Distance to nearest road (Dist.\ road)} Prediction of distance to the nearest road, derived from the GRIP4 dataset \citep{meijer2018global}. We calculated the distance-to-the-nearest road for every (10 m resolution) pixel in a 128 $\times$ 128 pixel patch, with a maximum radius of 50 km, and averaged to obtain the mean distance-to-road. 

\subsection{Sampling a land cover stratified dataset}\label{sec:sampling_lc_locations}
We created a new dataset of Earth embedding and land cover data for 10k randomly sampled, land cover (LC)-stratified locations across the globe. Further, we created a dataset of 10k randomly and spatially uniformly sampled locations across the Earth's landmass (excluding Antarctica). In the following, we detail the sampling strategy for the LC-stratified dataset and detail properties of both datasets.

We first created a dataset of 100k locations sampled uniformly from the globe. To do so, we sampled points randomly from a spherical surface and discarded all points below $56\degree$S latitude. We then retrieved one land cover of each location, by creating a 50 m square buffer around each point and averaging the 2024 DynamicWorld land cover values of those pixels (across pixels and time) \citep{brown2022dynamic}. (We extracted the land cover values of the centre pixel for the regression task.) Locations with no land cover data available (\textit{e.g.}, oceans) were discarded until 100k locations with land cover data were reached. 

DynamicWorld consists of 9 classes: \textit{water}, \textit{trees}, \textit{grass}, \textit{scrub and shrub}, \textit{bare}, \textit{flooded vegetation}, \textit{crops}, \textit{built}, and \textit{snow and ice}. The time-averaged DynamicWorld land cover data yields a value between 0 and 1 for each land cover class, which we refer to as `probabilities' for ease of notation, which corresponds to the time-averaged predicted probability $p_l(c)$ of each land cover class $c$, normalised to 1 across classes per location $l$: $\sum_c p_l(c) = 1$. This enabled us to calculate the distribution of land cover classes across locations in two ways: `winner-takes-all', where the max-probability labels are summed, and `weighted' where the probabilities are summed. This demonstrates that some classes rarely occur as majority class, despite their weighed abundance, because of temporal averaging (\textit{flooded vegetation}) or spatial averaging (\textit{built}). 
To assess the uniformity of the summed discrete (land cover class) distributions $p(c) = \frac{1}{L} \sum_{l=1}^Lp_l(c)$, we consider two metrics: the sampling efficiency coefficient $C_{\text{eff}} = \frac{\text{min}(p(c))}{\text{mean}(p(c))}\geq 0$ and entropy $H = -\sum_c p(c) \log \left( p(c) \right) > 0$. For a perfectly uniform distribution, $H = \log N_c$ (where $N_c = 9$ is the number of classes) and $C_\text{eff} = 1$, which both are the respective upper bounds.

\begin{figure}[b]
    \centering
    \includegraphics[width=\linewidth]{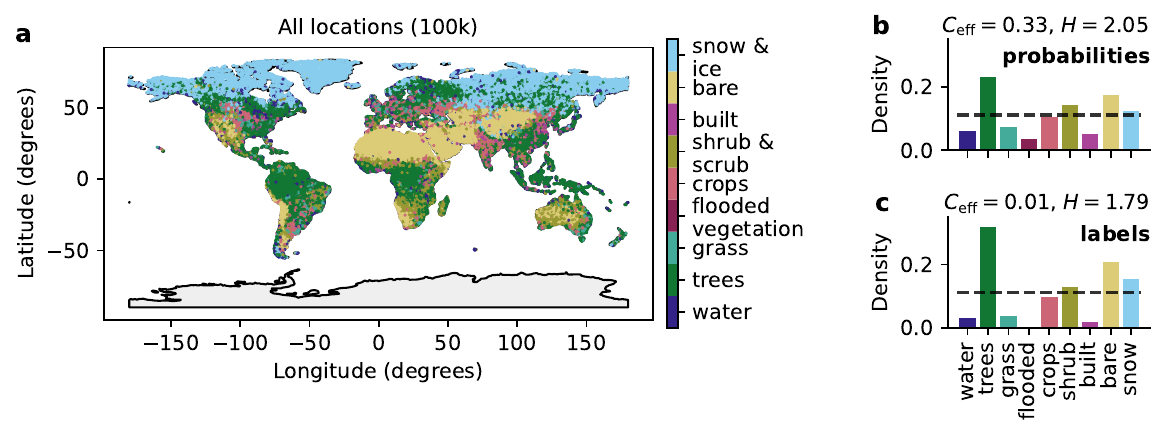}
    \caption{\textbf{Dataset of 100k randomly sampled points.} \textbf{a)} Points were randomly drawn from a 3D sphere, above 56$\degree$S where land cover data was available. \textbf{b-c)} Some land cover types, such as \textit{flooded vegetation} and \textit{built}, are rarely the majority class (c) and are better represented by summing the probabilities (b). }
    \label{fig:map_all}
\end{figure}

\begin{figure}
    \centering
    \includegraphics[width=\linewidth]{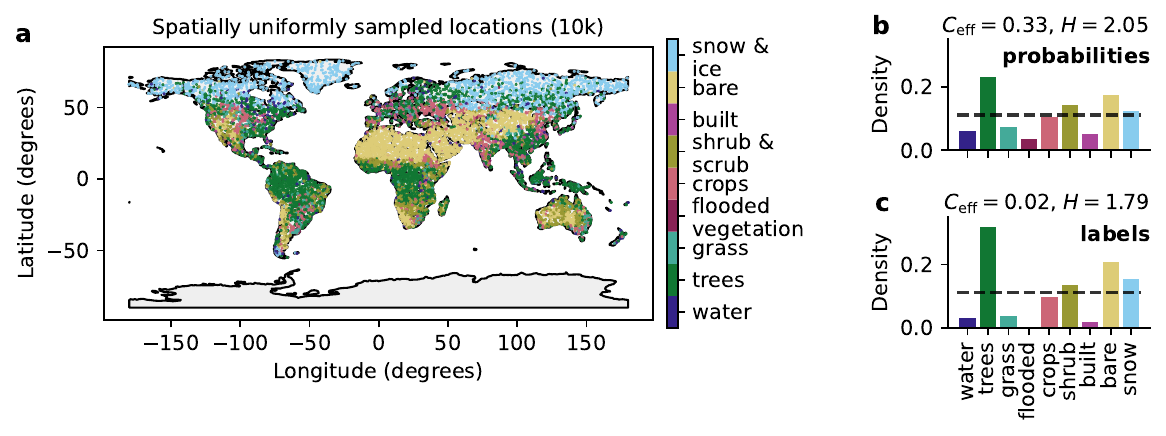}
    \caption{\textbf{Spatially uniform random sample of 10k locations.}}
    \label{fig:map_random}
\end{figure}

\begin{figure}
    \centering
    \includegraphics[width=\linewidth]{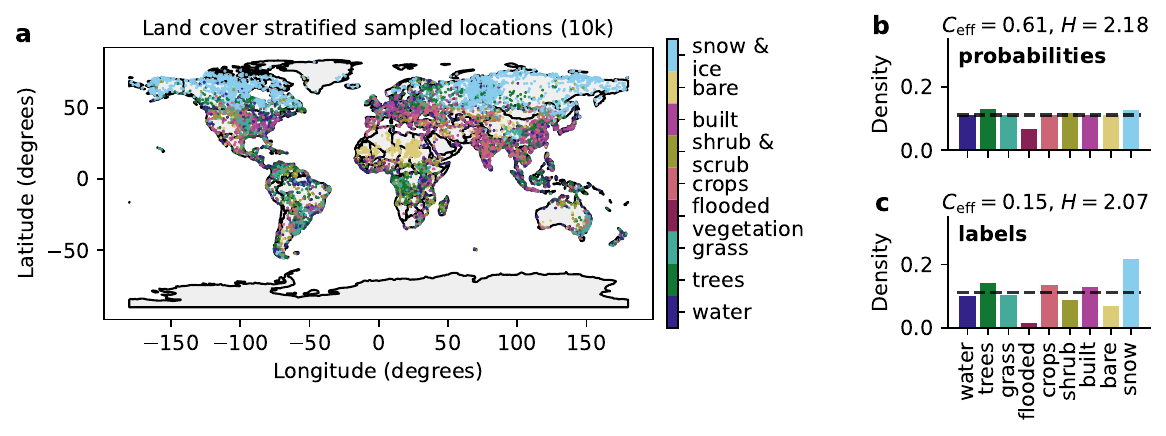}
    \caption{\textbf{Land cover stratified random sample of 10k locations.} The probabilities are balanced for all classes except \textit{flooded vegetation}. We confirmed that it is not possible to balance \textit{flooded vegetation} (by picking the 10k points with highest probability for this class), presumably due to the seasonal nature of this land cover class. See Fig \ref{fig:lc_maps} for the probabilities of each class across all points.}
    \label{fig:map_lc_strat}
\end{figure}

\begin{figure}
    \centering
    \includegraphics[width=0.4\linewidth]{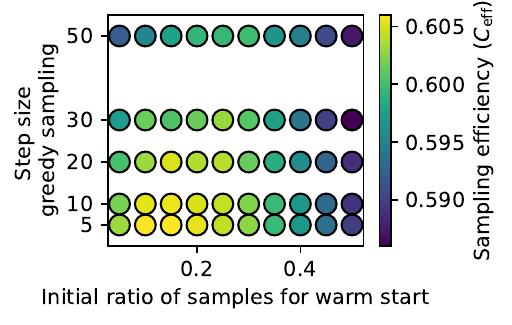}
    \caption{\textbf{Sampling efficiency was used to optimise the two hyperparameters of the greedy sampler.} A maximum of $C_\text{eff} = 0.6062$ was found for step size = 5 and initial ratio = 0.15.}
    \label{fig:sampler_optim}
\end{figure}

We created two `sample' datasets $D_S$ of 10k locations, sampled from the `population' dataset $D_P$ of 100k locations (Fig \ref{fig:map_all}): \textit{random}, where locations are sampled spatially uniformly randomly (Fig \ref{fig:map_random}, and \textit{LC-stratified}, where we used a greedy sampler to obtain a balanced distribution of land cover classes (Fig \ref{fig:map_lc_strat}). The greedy sampler is initiated with high-entropy locations (\textit{i.e.}, with mixed land cover), after which it iteratively samples locations that have high land cover probabilities for below-mean land cover classes (Algorithm \ref{alg:greedy-lc}). We decided to optimise for maximum sampling efficiency, and optimised the two hyperparameters of Algorithm \ref{alg:greedy-lc} (step size and initial ratio) for maximum sampling efficiency (Fig \ref{fig:sampler_optim}). 

\begin{algorithm}
\caption{Greedy sampler to obtain a land cover (LC)-stratified dataset of locations.}\label{alg:greedy-lc}
\begin{algorithmic}
\Require $n > 0$  \Comment{Desired sample size (10k).}
\Require $len(D_P) = N$ \Comment{Population dataset of 100k locations.}

\State $\texttt{step\_size}\gets 5$
\State $\texttt{initial\_ratio} \gets 0.15$ \\

\If{$\texttt{initial\_ratio} > 0$} \Comment{Initialise with the highest-entropy points.}
    \State $D_S \gets \textbf{sort}_H(D_P)$[:$(\texttt{initial\_ratio} * N)$]
    \State $D_P \gets D_P \setminus D_S$ \Comment{Disable sampling with replacement.}
\ElsIf{$\texttt{initial\_ratio} = 0$}
    \State $D_S \gets \{\}$
\EndIf \\

\While{$len(D_S) < n$}
    \State $\textbf{sample } c \sim \text{max}(\text{mean}(p) - p_c, 0)$  \Comment{Randomly pick one below-mean class, weighted.}
    \State $D_S \gets D_S \cup \textbf{sort}_{p_c}(D_P)$[:$\texttt{step\_size}$]  \Comment{Add locations with most LC of sampled class.}
    \State $D_P \gets D_P \setminus D_S$ \Comment{Disable sampling with replacement.}
\EndWhile \\

\Return $D_S$
\end{algorithmic}
\end{algorithm}

\begin{figure}
    \centering
    \includegraphics[width=\linewidth]{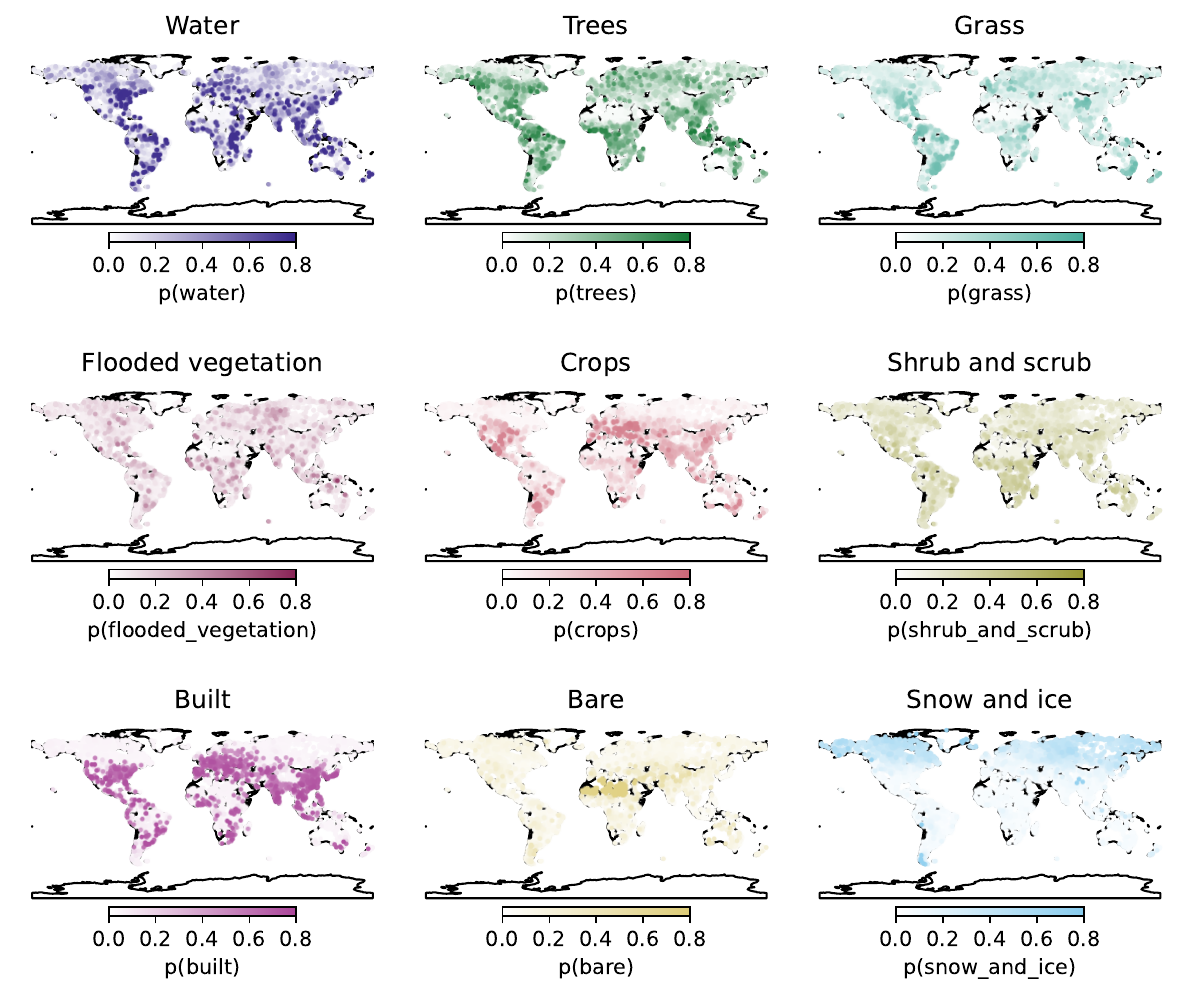}
    \caption{\textbf{Distribution of Dynamic World land cover data for the 8,829 sampled land cover stratified locations.}}
    \label{fig:lc_maps}
\end{figure}

\subsection{Supplementary results}

\input{tables/r2_scores_main_tasks_biomass}

\begin{figure}
    \centering
    \includegraphics[width=\linewidth]{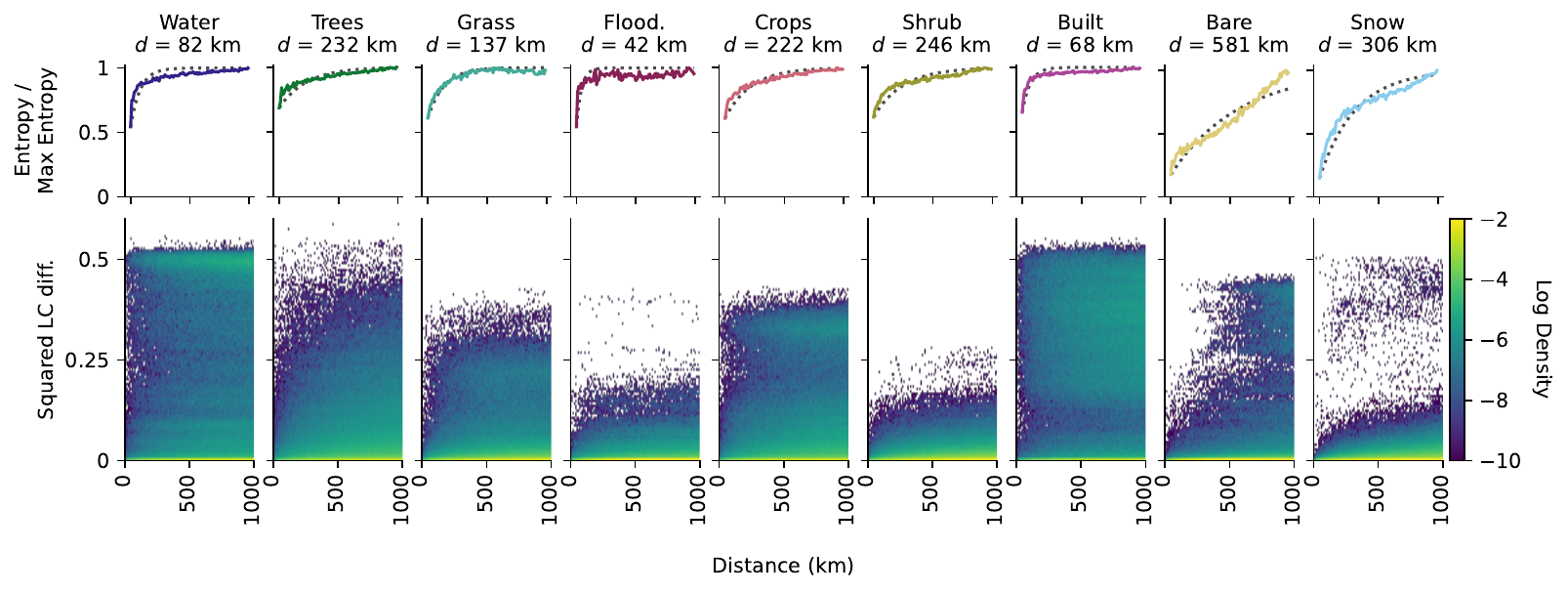}
    \caption{\textbf{Land cover classes are differentiated by their spatial scales.} Bottom row: weighed density plots of L2 differences between land cover class probabilities for all pairs of locations versus distance ($\leq 1000$ km). Top row: Entropy per distance bin (solid line) and exponential fit (dotted line) per land cover class.}
    \label{fig:spatial_entropy_lc}
\end{figure}

%% file: tables/dataset_overview.tex
\begin{table}[]
\caption{\textbf{Overview of downstream tasks.}}
\label{tab:overview_tasks}
\begin{adjustbox}{max width=\textwidth}
\begin{tabular}{llllll}
\toprule
                   & Task                      & \# target variables & \# locations & Source         & License \\ \midrule
Crop type          & Multiclass classification & 18         & 3,596         & Crop Harvest  & CC-BY-SA 4.0 \\
Biomass            & Univariate regression     & 1          & 4,566         & MMEarth-Bench & CC-BY \\
Land cover         & Multivariate regression   & 9          & 8,829         & Dynamic World & CC-BY 4.0  \\
Bioclimatic        & Multivariate regression   & 19         & 8,901         & WorldClim BIO & CC-BY-SA 4.0  \\
Population density & Univariate regression     & 1          & 8,829         & WorldPop      & CC BY 4.0  \\
Distance to road   & Univariate regression     & 1          & 8,901         & GRIP4         & CC-0 \\    \bottomrule
\end{tabular}
\end{adjustbox}
\end{table}

%% file: tables/models_overview.tex

\begin{table}[t]
\centering
\caption{\textbf{Overview of pre-computed Earth embeddings that were evaluated.}}
\label{tab:overview_models}
\begin{adjustbox}{max width = \textwidth}
\begin{tabular}{lllll}
\toprule 
           & Type               & Dim. & Source   & License  \\ \midrule
alphaearth & Pixel embedding    & 64         &\citet{brown2025alphaearth}  & CC-BY 4.0   \\
tessera    & Pixel embedding    & 128        & \citet{feng2025tessera}       & CC-0       \\
geoclip    & Location embedding & 256        & \citet{vivanco2023geoclip} & MIT   \\
satclip    & Location embedding & 512        & \citet{klemmer2025satclip}   & MIT        \\ \bottomrule
\end{tabular}
\end{adjustbox}
\end{table}

%% file: tables/r2_scores_main_tasks.tex
\begin{table}[ht]
\centering
\caption{\textbf{Fused embeddings typically outperform single embeddings, and the best combination is task-dependent}. $R^2$ [\%] scores for regression tasks (ridge regression), accuracy [\%] for the classification task (logistic regression). Best score for each task is highlighted in bold (mean $\pm$ SEM), for single-embedding, multi-embedding and all embeddings (if better than previous).}
\label{tab:r2_scores_main_tasks}
\begin{adjustbox}{max width = \textwidth}
\begin{tabular}{lllllll}
\toprule
Embeddings & Crops & Biomass & Land cover & Bioclimatic & Pop. & Dist. road \\
\midrule
alphaearth & 60.0 ± 0.8 & \textbf{67.1 ± 2.9} & \textbf{71.1 ± 0.4} & \textbf{83.5 ± 0.4} & \textbf{26.7 ± 2.2} & 49.0 ± 0.6 \\
tessera & 56.5 ± 0.5 & 52.3 ± 2.4 & 70.6 ± 0.4 & 77.4 ± 0.5 & 18.0 ± 3.0 & \textbf{51.2 ± 0.6} \\
geoclip & \textbf{61.6 ± 0.7} & 30.8 ± 2.4 & 30.0 ± 0.6 & 80.4 ± 0.4 & -3.7 ± 3.9 & 51.0 ± 0.8 \\
satclip & 45.9 ± 0.5 & 34.5 ± 2.2 & 33.1 ± 0.6 & 74.9 ± 0.4 & 3.2 ± 1.8 & 49.2 ± 0.5 \\ \midrule
alphaearth + tessera & 57.5 ± 0.7 & 67.1 ± 2.9 & \textbf{76.7 ± 0.3} & 85.5 ± 0.4 & 26.0 ± 2.7 & 54.0 ± 0.6 \\
alphaearth + geoclip & \textbf{63.4 ± 0.7} & 63.5 ± 2.9 & 71.4 ± 0.4 & 86.7 ± 0.4 & 15.8 ± 3.8 & 55.0 ± 0.7 \\
alphaearth + satclip & 60.6 ± 0.6 & 66.7 ± 2.9 & 72.7 ± 0.4 & \textbf{86.9 ± 0.3} & 23.8 ± 2.6 & 55.2 ± 0.5 \\
tessera + geoclip & 62.3 ± 0.7 & 50.8 ± 2.7 & 70.4 ± 0.4 & 84.4 ± 0.4 & 8.5 ± 4.5 & \textbf{56.3 ± 0.6} \\
tessera + satclip & 57.5 ± 0.6 & 54.7 ± 2.6 & 71.6 ± 0.3 & 83.2 ± 0.4 & 14.3 ± 3.6 & 56.0 ± 0.6 \\
geoclip + satclip & 61.7 ± 0.6 & 32.5 ± 2.7 & 33.7 ± 0.7 & 84.9 ± 0.4 & -8.0 ± 4.9 & 55.6 ± 0.7 \\ \midrule
All GFMs & 62.5 ± 0.6 & 60.5 ± 3.1 & 76.2 ± 0.4 & \textbf{89.2 ± 0.3} & 10.8 ± 4.9 & \textbf{59.0 ± 0.6} \\
\bottomrule
\end{tabular}
\end{adjustbox}
\end{table}

%% file: tables/compl_r2_scores_main_tasks.tex
\begin{table}[ht]
\caption{\textbf{Complementarity of embeddings for downstream tasks}. Complementarity was calculated from $R^2$ scores for the regression tasks and accuracy for the classification task. Positive values are highlighted in bold and indicate positive complementarity.  ***: $p < 0.001$; **: $p < 0.01$; *: $p < 0.05$.}
\label{tab:compl_r2_scores_main_tasks}
\centering
\begin{tabular}{lllllll}
\toprule
Embeddings & Crops & Biomass & Land cover & Bioclimatic & Pop. & Dist. road \\
\midrule
alphaearth + tessera & -0.06 & 0.00 & \textbf{0.19***} & \textbf{0.12***} & -0.01 & \textbf{0.06***} \\
alphaearth + geoclip & \textbf{0.05**} & -0.11 & \textbf{0.01**} & \textbf{0.19***} & -0.15 & \textbf{0.08***} \\
alphaearth + satclip & 0.01 & -0.01 & \textbf{0.05***} & \textbf{0.20***} & -0.04 & \textbf{0.12***} \\
tessera + geoclip & 0.02 & -0.03 & -0.01 & \textbf{0.20***} & -0.12 & \textbf{0.10***} \\
tessera + satclip & \textbf{0.02**} & \textbf{0.05**} & \textbf{0.04***} & \textbf{0.26***} & -0.04 & \textbf{0.10***} \\
geoclip + satclip & 0.00 & -0.03 & \textbf{0.01*} & \textbf{0.23***} & -0.12 & \textbf{0.10***} \\ \midrule
All GFMs & 0.02 & -0.20 & \textbf{0.18***} & \textbf{0.35***} & -0.22 & \textbf{0.16***} \\
\bottomrule
\end{tabular}
\end{table}

%% file: tables/r2_scores_lc_classes.tex
\begin{table}[ht]
\caption{\textbf{Class-specific scores for land cover regression}. $R^2$ [\%] scores per class (ridge regression). Best score for each task is highlighted in bold (mean $\pm$ SEM), for single-embedding, multi-embedding and all embeddings (if better than previous).}
\label{tab:r2_scores_lc_classes}
\begin{adjustbox}{max width = \textwidth}
\begin{tabular}{llllllllll}
\toprule
Embeddings & \textit{Water} & \textit{Trees} & \textit{Grass} & \textit{Flood.} & \textit{Crops} & \textit{Shrub} & \textit{Built} & \textit{Bare} & \textit{Snow} \\
\midrule
alphaearth & \textbf{90.7 ± 0.4} & \textbf{70.6 ± 0.7} & 64.2 ± 0.9 & 34.2 ± 0.7 & 66.5 ± 0.6 & 54.9 ± 0.7 & \textbf{80.9 ± 0.6} & 91.6 ± 0.4 & 86.0 ± 0.5 \\
tessera & 85.5 ± 0.8 & 66.5 ± 0.9 & \textbf{68.9 ± 0.9} & \textbf{36.6 ± 0.7} & \textbf{68.0 ± 0.6} & \textbf{55.9 ± 0.7} & 73.1 ± 0.6 & \textbf{91.8 ± 0.4} & \textbf{87.9 ± 0.5} \\
geoclip & 3.5 ± 1.6 & 13.7 ± 0.8 & 24.6 ± 1.2 & 6.5 ± 0.8 & 24.3 ± 1.0 & 16.9 ± 1.1 & 18.9 ± 1.0 & 76.3 ± 1.2 & 83.1 ± 0.7 \\
satclip & 7.6 ± 1.4 & 17.8 ± 1.0 & 30.5 ± 1.3 & 12.9 ± 0.7 & 30.6 ± 1.0 & 22.4 ± 1.3 & 19.1 ± 1.0 & 75.0 ± 1.0 & 80.1 ± 0.7 \\ \midrule
alphaearth + tessera & \textbf{92.3 ± 0.4} & \textbf{74.9 ± 0.7} & \textbf{74.4 ± 0.8} & \textbf{43.3 ± 0.8} & \textbf{74.9 ± 0.5} & \textbf{63.1 ± 0.7} & \textbf{84.1 ± 0.5} & \textbf{93.5 ± 0.4} & \textbf{89.0 ± 0.4} \\
alphaearth + geoclip & 90.6 ± 0.4 & 70.0 ± 0.7 & 64.9 ± 0.9 & 33.5 ± 0.9 & 67.6 ± 0.6 & 55.0 ± 0.7 & 80.9 ± 0.6 & 91.8 ± 0.4 & 87.3 ± 0.5 \\
alphaearth + satclip & 91.1 ± 0.4 & 71.5 ± 0.7 & 66.6 ± 0.9 & 36.0 ± 0.7 & 69.2 ± 0.5 & 57.8 ± 0.8 & 81.5 ± 0.6 & 92.2 ± 0.4 & 87.7 ± 0.5 \\
tessera + geoclip & 85.1 ± 0.8 & 65.8 ± 0.9 & 69.3 ± 0.9 & 34.9 ± 1.0 & 68.5 ± 0.6 & 55.9 ± 0.8 & 72.3 ± 0.6 & 91.8 ± 0.4 & 88.7 ± 0.4 \\
tessera + satclip & 85.7 ± 0.8 & 67.1 ± 0.9 & 70.8 ± 0.8 & 37.7 ± 0.8 & 70.0 ± 0.6 & 58.1 ± 0.7 & 72.9 ± 0.6 & 92.2 ± 0.4 & 89.0 ± 0.4 \\
geoclip + satclip & 5.6 ± 2.0 & 16.1 ± 1.1 & 31.1 ± 1.5 & 9.6 ± 1.0 & 30.0 ± 1.1 & 21.5 ± 1.4 & 19.3 ± 1.1 & 82.3 ± 1.0 & 84.9 ± 0.7 \\ \midrule
All GFMs & 91.9 ± 0.4 & 73.8 ± 0.8 & 74.4 ± 0.8 & 40.4 ± 0.9 & \textbf{75.1 ± 0.5} & 62.6 ± 0.7 & 83.4 ± 0.5 & \textbf{93.5 ± 0.4} & \textbf{89.9 ± 0.4} \\
\bottomrule
\end{tabular}
\end{adjustbox}
\end{table}

%% file: tables/compl_r2_scores_lc_classes.tex
\begin{table}[ht]
\caption{\textbf{Complementarity of embeddings for land cover regression}. Complementarity was calculated from $R^2$ scores (ridge regression). Positive values are highlighted in bold and indicate positive complementarity.  ***: $p < 0.001$; **: $p < 0.01$; *: $p < 0.05$.}
\label{tab:compl_r2_scores_lc_classes}
\begin{adjustbox}{max width = \textwidth}
\begin{tabular}{llllllllll}
\toprule
Embeddings & \textit{Water} & \textit{Trees} & \textit{Grass} & \textit{Flood.} & \textit{Crops} & \textit{Shrub} & \textit{Built} & \textit{Bare} & \textit{Snow} \\
\midrule
alphaearth + tessera & \textbf{0.17***} & \textbf{0.15***} & \textbf{0.18***} & \textbf{0.11***} & \textbf{0.22***} & \textbf{0.16***} & \textbf{0.17***} & \textbf{0.21***} & \textbf{0.09***} \\
alphaearth + geoclip & -0.01 & -0.02 & \textbf{0.02*} & -0.01 & \textbf{0.03**} & 0.00 & 0.00 & \textbf{0.02*} & \textbf{0.10***} \\
alphaearth + satclip & \textbf{0.04***} & \textbf{0.03***} & \textbf{0.07***} & \textbf{0.03**} & \textbf{0.08***} & \textbf{0.06***} & \textbf{0.03***} & \textbf{0.08***} & \textbf{0.13***} \\
tessera + geoclip & -0.03 & -0.02 & 0.01 & -0.03 & 0.02 & 0.00 & -0.03 & 0.01 & \textbf{0.07***} \\
tessera + satclip & 0.01 & \textbf{0.02**} & \textbf{0.06***} & \textbf{0.02*} & \textbf{0.06***} & \textbf{0.05***} & 0.00 & \textbf{0.04***} & \textbf{0.09***} \\
geoclip + satclip & -0.02 & -0.02 & 0.01 & -0.04 & -0.01 & -0.01 & 0.00 & \textbf{0.25***} & \textbf{0.10***} \\ \midrule
All GFMs & \textbf{0.13***} & \textbf{0.11***} & \textbf{0.17***} & \textbf{0.06***} & \textbf{0.22***} & \textbf{0.15***} & \textbf{0.13***} & \textbf{0.21***} & \textbf{0.16***} \\
\bottomrule
\end{tabular}
\end{adjustbox}
\end{table}

%% file: tables/spatial_relationship_r2.tex
\caption{Spearman correlation between the spatial scale of each land cover class $d$ and the $R^2$ prediction score for that land cover class from single-GFM embeddings.}
\label{tab:spatial_r2}
\begin{tabular}{ll}
\toprule
Embeddings & $\rho(R^2, d)$ \\
\midrule
alphaearth & 0.40 \\
tessera & 0.45 \\
geoclip & $\textbf{0.68*}$ \\
satclip & $\textbf{0.80*}$ \\
\bottomrule
\end{tabular}

%% file: tables/spatial_relationship_compl.tex
\caption{Spearman correlation between the spatial scale of each land cover class $d$ and the complementarity score of an embedding pair. }
\label{tab:spatial_compl}
\begin{tabular}{ll}
\toprule
Embeddings & $\rho(C^\text{LC}, d)$ \\
\midrule
alphaearth + tessera & 0.17 \\
alphaearth + geoclip & 0.60 \\
alphaearth + satclip & $\textbf{0.82*}$ \\
tessera + geoclip & $\textbf{0.70*}$ \\
tessera + satclip & $\textbf{0.68*}$ \\
geoclip + satclip & $\textbf{0.63*}$ \\ \midrule
All GFMs & $\textbf{0.68*}$ \\
\bottomrule
\end{tabular}

%% file: tables/r2_scores_main_tasks_biomass.tex
\begin{table}[ht]
\caption{\textbf{Fused embeddings typically outperform single embeddings, and the best combination is task-dependent}. $R^2$ [\%] scores for regression tasks (ridge regression), accuracy [\%] for the classification task (logistic regression). Best score for each task is highlighted in bold (mean $\pm$ SEM), for single-embedding, multi-embedding and all embeddings (if better than previous).}
\label{tab:r2_scores_main_tasks_biomass}
\begin{tabular}{lll}
\toprule
Embeddings & Biomass (centre-pixel) & Biomass (mean) \\
\midrule
alphaearth & \textbf{49.9 ± 2.8} & \textbf{67.1 ± 2.9} \\
tessera & 37.3 ± 2.3 & 52.3 ± 2.4 \\
geoclip & 11.4 ± 1.5 & 30.8 ± 2.4 \\
satclip & 17.4 ± 1.7 & 34.5 ± 2.2 \\
alphaearth + tessera & 49.0 ± 2.8 & 67.1 ± 2.9 \\
alphaearth + geoclip & 42.0 ± 2.4 & 63.5 ± 2.9 \\
alphaearth + satclip & 47.2 ± 2.8 & 66.7 ± 2.9 \\
tessera + geoclip & 30.5 ± 2.0 & 50.8 ± 2.7 \\
tessera + satclip & 36.1 ± 2.3 & 54.7 ± 2.6 \\
geoclip + satclip & 9.1 ± 1.6 & 32.5 ± 2.7 \\
All GFMs & 36.6 ± 2.4 & 60.5 ± 3.1 \\
\bottomrule
\end{tabular}
\end{table}